\documentclass[runningheads]{llncs}

\usepackage{eccv}

\usepackage{eccvabbrv}

\usepackage{graphicx}
\usepackage{booktabs}
\usepackage{enumerate}
\usepackage{algorithm}
\usepackage{algorithmic}
\usepackage{colortbl}
\usepackage{makecell}
\usepackage{multirow}
\usepackage{wrapfig}
\usepackage{array}
\usepackage{amsmath,bm}
\usepackage[marginal]{footmisc}

\usepackage[accsupp]{axessibility}  

\usepackage[pagebackref,breaklinks,colorlinks,citecolor=eccvblue]{hyperref}

\begin{document}

\title{Knowledge Condensation and Reasoning for Knowledge-based VQA}

\titlerunning{Knowledge Condensation and Reasoning for Knowledge-based VQA}

\author{Dongze Hao\inst{1,2}\thanks{Work done during internship at Kuaishou.} \and
Jian Jia\inst{3} \and Longteng Guo\inst{1} \and Qunbo Wang\inst{1} \and Te Yang\inst{1,2} \and Yan Li\inst{3} \and Yanhua Cheng\inst{3} \and Bo Wang\inst{3} \and Quan Chen\inst{3} \and Han Li\inst{3} \and Jing Liu\inst{1,2}\thanks{Corresponding author.}}

\authorrunning{Dongze Hao et al.}

\institute{Laboratory of Cognition and Decision Intelligence for Complex Systems, Institute of Automation, Chinese Academy of Sciences, 100190, Beijing, China \and School of Artificial Intelligence, University of Chinese Academy of Sciences, 101408, Beijing, China \and Kuaishou, Beijing, China}
\maketitle
\footnote{First Author and Second Author contribute equally to this work.\\}
\begin{abstract}

Knowledge-based visual question answering (KB-VQA) is a challenging task, which requires the model to leverage external knowledge for comprehending and answering questions grounded in visual content. Recent studies retrieve the knowledge passages from external knowledge bases and then use them to answer questions. However, these retrieved knowledge passages often contain irrelevant or noisy information, which limits the performance of the model. To address the challenge, we propose two synergistic models: \textbf{Knowledge Condensation} model and \textbf{Knowledge Reasoning} model.
We condense the retrieved knowledge passages from two perspectives. First, we leverage the multimodal perception and reasoning ability of the visual-language models to distill concise knowledge concepts from retrieved lengthy passages, ensuring relevance to both the visual content and the question. Second, we leverage the text comprehension ability of the large language models to summarize and condense the passages into the knowledge essence which helps answer the question. These two types of condensed knowledge are then seamlessly integrated into our Knowledge Reasoning model, which judiciously navigates through the amalgamated information to arrive at the conclusive answer.
Extensive experiments validate the superiority of the proposed method. Compared to previous methods, our method achieves state-of-the-art performance on knowledge-based VQA datasets (\textbf{65.1}\% on OK-VQA and \textbf{60.1}\% on A-OKVQA) without resorting to the knowledge produced by GPT-3 (175B).


    
\end{abstract}

\section{Introduction}
\label{sec:intro}

Visual question answering (VQA) \cite{antol2015vqa} is a task requiring the AI model to generate answers to questions grounded in the visual content presented within an image. Initial datasets \cite{goyal2017making, hudson2019gqa} in VQA predominantly concentrated on the facets of visual perception and recognition. 
However, in real-world scenarios, it is often imperative for humans to assimilate world knowledge when making decisions based on visual information. To address this complexity, the task of knowledge-based visual question answering (KB-VQA) \cite{marino2019ok} is proposed. This paradigm shift encourages the development of models that are capable of utilizing external knowledge sources to respond to open-ended questions that are visually anchored, thereby simulating a more realistic and human-like understanding and interaction with the visual world.

\begin{figure}[t]
  \centering
   \includegraphics[width=1.0\linewidth]{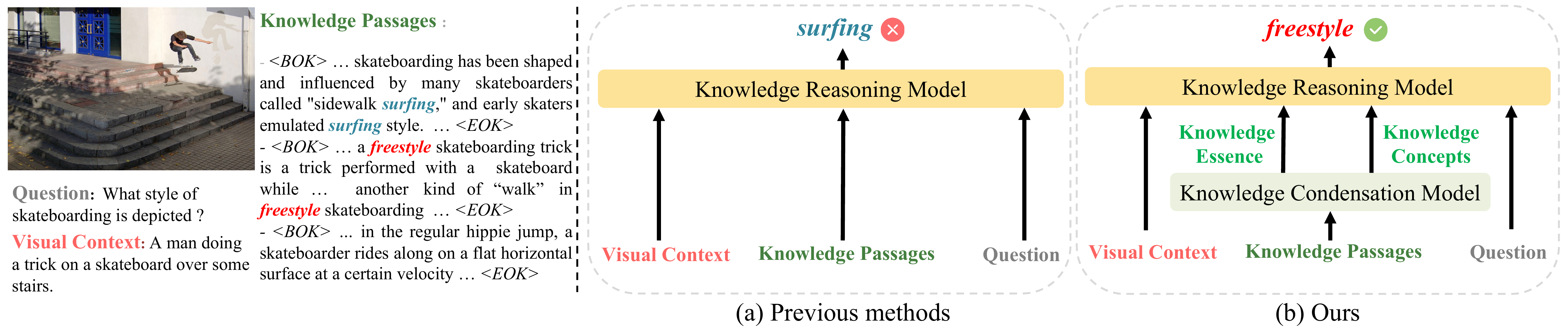}
   \caption{ {\bf Comparison with previous methods}. (a) Previous methods \cite{gui2021kat, lin2022revive, lin2022retrieval} convert the images to visual contexts (captions) and send them to the LLM along with the questions and retrieved knowledge passages to predict the answers. Due to the retrieved knowledge passages contain many noisy information, they mislead the model to predict the wrong answer "surfing". (b) We leverage the trainable VLM and frozen LLM to condense lengthy knowledge passages into concise \textbf{knowledge concepts} and \textbf{knowledge essence} to mitigate the interference of noisy information. With the condensed knowledge, our knowledge reasoning model generates the right answer.}
   \label{fig:intro}
   \vspace{-1em}
\end{figure}

Existing state-of-the-art methods \cite{gui2021kat, gao2022transform, lin2022revive} mostly follow a "retrieve and generate" paradigm to take the advantage of external knowledge to solve the task of KB-VQA (shown in Fig.~\ref{fig:intro} (a)). They retrieve the knowledge passages from explicit knowledge bases (ConceptNet \cite{speer2017conceptnet}, Wikipedia \cite{vrandevcic2014wikidata}, and Google Search Corpus \cite{luo2021weakly}) and leverage the pre-trained language model to generate the final answer. Except for the explicit knowledge, some methods \cite{gui2021kat, lin2022revive, hu2022promptcap, shao2023prompting} take GPT-3 \cite{brown2020language}, a large language model trained on huge amounts of corpus and demonstrated incomparable power over common sense reasoning, as the implicit KBs to generate related knowledge and answers. These methods focus on augmenting various knowledge bases to improve the KB-VQA model performance. Nevertheless, there is little concern about how to utilize explicit retrieved knowledge bases more effectively. 
We argue that using explicit knowledge passages retrieved by DPR \cite{karpukhin2020dense} introduces irrelevant information, which misleads and distracts the language model from reasoning the right answer. Given the limited input capacity of language models, it is particularly important to filter condensed knowledge content from verbose passages and information.

To efficiently utilize the retrieved knowledge passages, we propose the knowledge condensation model and the knowledge reasoning model to extract key information from the knowledge and reason over the knowledge for answering the question, respectively. 
Specifically, we first utilize the multimodal perception and reasoning ability of the visual-language model to condense the retrieved knowledge passages into concise knowledge concepts that are tightly related to visual and question information. 
Then we take advantage of the text comprehension ability of the large language model to extract key knowledge essence which helps to answer the question. Finally,
we integrate the concise and informative knowledge concepts and essence as well as other implicit knowledge into the knowledge reasoning model, a unified encoder-decoder language model, to predict the final answer.

Our contributions are summarized as follows:

\begin{enumerate}[(a)]

\item We establish an effective knowledge condensation and reasoning framework for knowledge-based visual question answering (KB-VQA), which extracts informative knowledge from lengthy passages of knowledge bases and avoids the influence of noisy information.
\item We propose two synergistic models: the knowledge condensation model and the knowledge reasoning model. The knowledge condensation models utilize a visual language model and a large language model to compress lengthy passages into distilled knowledge concepts and essence. The knowledge reasoning model integrates knowledge concepts and essence into a pre-trained large language model to reason out the final answer.

\item We achieve the state-of-the-art performance 65.1\% on OK-VQA and 60.1\% on A-OKVQA, surpassing SOTA methods by 3.0\% and 0.5\% respectively, without utilizing the knowledge generated by GPT-3 (175B).
\end{enumerate}

\section{Related work}
\label{sec:relatedwork}
\noindent\textbf{Visual Question Answering (VQA).} The traditional VQA task requires the AI model to answer the question about the image contents. Early studies \cite{yu2018beyond, anderson2018bottom} focus on the cross-modality fusion of visual features and textual features to predict the answer. Recently, visual-language models \cite{wang2022ofa,wang2022image,li2023blip, alayrac2022flamingo} based on the pretrain-finetune paradigm have achieved promising performances on conventional VQA benchmarks such as VQA v2 \cite{goyal2017making}. 

\vspace{0.5em}
\noindent\textbf{Knowledge-based VQA.} Knowledge-based VQA requires the AI model to answer visual-related questions utilizing external knowledge. Early studies \cite{wang2017fvqa, wang2015explicit} retrieve the knowledge from fixed knowledge bases and utilize the attention-based model to predict answers. Recently, OK-VQA \cite{marino2019ok} is proposed to encourage the model to utilize various world knowledge to answer questions instead of providing fixed knowledge bases. To solve the problem, researchers utilize the knowledge from various external knowledge bases, such as ConceptNet \cite{speer2017conceptnet}, Wikipedia \cite{vrandevcic2014wikidata}, Google Search Corpus \cite{luo2021weakly}, and GPT-3. For example, ConceptBERT \cite{garderes2020conceptbert} uses transformer to integrate the graph embeddings from ConceptNet with visual-language features to predict the answer. MAVEx \cite{wu2022multi} retrieves the knowledge from Wikipedia, ConceptNet and uses a pretrained visual-language model to generate the answer. TRiG \cite{gao2022transform} transforms the image into texts and uses DPR to retrieve relevant knowledge from Wikipedia, then utilizes a large language model to perform open-ended answer generation. Similar to TRiG, RA-VQA \cite{lin2022retrieval} uses DPR \cite{karpukhin2020dense} to retrieve the knowledge from Google Search Corpus and train the retriever and the answer generator simultaneously, which improves the performance of DPR and the final answer prediction performance. PICa \cite{yang2022empirical} prompts GPT-3 to predict the answer in an in-context learning pipeline. Prophet \cite{shao2023prompting} introduces the answer candidates to enhance the reasoning ability of GPT-3 similar to PICa. Different from prompt-based methods, KAT \cite{gui2021kat} and REVIVE \cite{lin2022revive} utilize GPT-3 \cite{brown2020language} to generate answer candidates and corresponding evidence and integrate them into a language model to predict the answer. 

\vspace{0.5em}
\noindent\textbf{Large Visual-Language Model.} Recently, large visual-language models \cite{li2023blip, alayrac2022flamingo} have demonstrated impressive performance on various visual-language tasks. The structure of these models usually consists of a frozen visual encoder \cite{radford2021learning}, a visual extractor \cite{li2023blip}, and a large language model \cite{chung2022scaling, zhang2022opt, vicuna}. They are first pretrained on large visual-language collections to inject visual information into the LLM. Then they finetune the model on various downstream tasks.
In this paper, we utilize the implicit reasoning of the language model to condense the knowledge guided by question and the image.

\vspace{0.5em}
\noindent\textbf{Large Language Model.} Recent autoregressive language models \cite{brown2020language,zhang2022opt, touvron2023llama, vicuna, touvron2023llama} have demonstrated strong language comprehension capabilities and instruction-following abilities. In this paper, we prompt the large language model to extract the key information which is related to the question. 

Compared with previous KB-VQA methods, our method focus on the effective usage of the retrieved knowledge. We propose to condense the lengthy knowledge passages into concise knowledge concepts and essence to alleviate the interference of extraneous information in the retrieved knowledge. With the condensed knowledge and other implicit knowledge, we use the knowledge reasoning model to reason over on various information to generate the answer for solving the task of KB-VQA.

\section{Method}
\label{sec:method}

The proposed method consists of two models: the knowledge condensation model and the knowledge reasoning model. We first use the knowledge condensation model to extract key information from noisy retrieved knowledge passages. Then we use the knowledge reasoning model to utilize various information to answer the question. Two models are detailed in Sec.~\ref{kc} and Sec.~\ref{kr}.

\begin{figure}[t]
  \centering
   \includegraphics[width=1.0\linewidth]{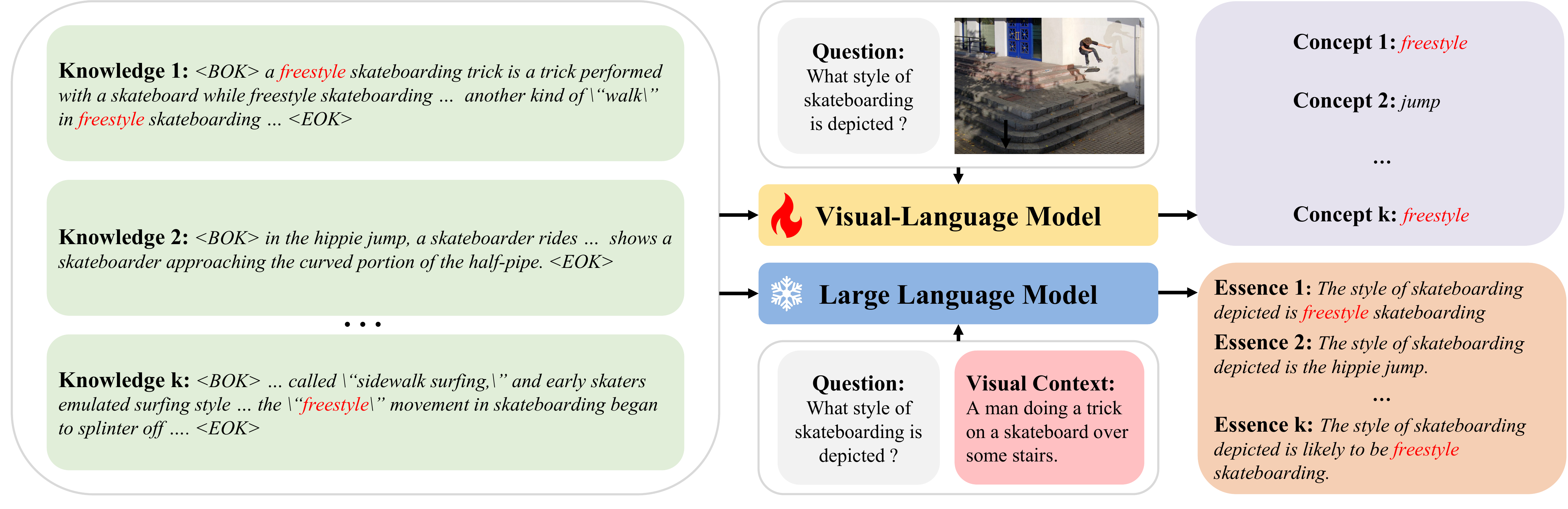}
   \caption{\textbf{The structure of the knowledge condensation model.} The knowledge condensation model consists of a visual-language model (VLM) and a large-language model (LLM). The VLM takes the image, question, and each retrieved passage as inputs and is trained by the supervision of the ground-truth answer. By utilizing the multimodal perception reasoning ability of VLM, each knowledge is condensed into the knowledge concept. The LLM takes the visual context, question, and each retrieved passage as inputs, we directly prompt LLM to condense each knowledge passage into the knowledge essence.}
   \label{fig:framework}
   \vspace{-1em}
\end{figure}

\subsection{Knowledge Retrieval}

Following the common settings \cite{gao2022transform,lin2022retrieval}, we adopt the Google Search Corpus (GSP) \cite{luo2021weakly} as the external knowledge base for OK-VQA. For A-OKVQA, we adopt the Wikidata \cite{vrandevcic2014wikidata} as the external knowledge base.
We transform the image into raw texts composed of captions, objects, attributes, and OCR (Optical Character Recognition). DPR is adopted \cite{karpukhin2020dense} as the knowledge retriever to index Top-K related knowledge passages $\bm{P} = \{P^1, P^2, ..., P^K\}$ from the knowledge base according to the raw texts like previous methods \cite{luo2021weakly, lin2022retrieval}.

\subsection{Knowledge Condensation Model}
\label{kc}
Given the retrieved knowledge passages, previous methods \cite{lin2022retrieval, gui2021kat, lin2022revive} directly integrate them into the model for answer generation. Although these studies have proven that the retrieved knowledge passages are beneficial to the model to get the right answer, these lengthy passages inevitably contain irrelevant and distractive information that impedes the answer reasoning process. We wish to extract the key information from the retrieved knowledge passages to help the model mitigate the influence of irrelevant information. To achieve this, we leverage a visual-language model and a large language model to condense lengthy passages into refined knowledge concepts and knowledge essence which help to answer the question.




\noindent\textbf{Visual-Language Model.}
We adopt BLIP2 \cite{li2023blip} as the visual-language model in this paper. BLIP2 is a recent state-of-the-art visual-language model, which has demonstrate strong multimodal perception and reasoning ability. So we use it as a condensation model to extract useful information from the retrieved knowledge. It consists of a frozen image encoder \cite{dosovitskiy2020image, fang2023eva}, a trainable Q-Former and a frozen pre-trained language model \cite{chung2022scaling, zhang2022opt}. Q-Former is a transformer-like architecture \cite{vaswani2017attention} which bridges image encoder and LLM. The image first fed into the image encoder to extract visual features $v$. Then the visual features $v$ and learnable queries $q$ are fed into Q-Former to output fixed-length visual features $h$. Finally, new visual features $h$ and the text embeddings $t$ are fed into the LLM to generate new texts. 

Previous works \cite{ouyang2022training, gui2021kat, lin2022revive, shao2023prompting} prompt GPT-3 \cite{brown2020language} or train a classified model to generated answer candidates as the concise knowledge, which helps the model to answer the question. Inspired by this, we use the ground-truth answer as the supervisory signal to train the visual-language model to predict the answer according to each retrieved knowledge passage. By utilizing the multimodal perception and reasoning ability of VLM, even though the retrieved knowledge contains irrelevant information, the model can filter the noisy information and give concise knowledge concepts which are related to the visual and question information.

Given the $i$-th sample $\{I_{i}, Q_{i}, P_{i}^{j}\}$, where $I$ denotes the image, $Q$ denotes the question and $P^{j}$ denotes the $j$-th knowledge passage ordered by relevance of DPR \cite{karpukhin2020dense}, we take triplets $\{I_{i}, Q_{i}, P_{i}^{j}\}$ of all samples as training data to train the condensation model under the supervision of the ground truth answer $A_{i}$:
\begin{equation}
    \label{eq:kc_loss}
    \bm{L}_{kc} = - \sum_{i=1}^N \sum_{j=1}^{K} \log \mathcal{M}_{kc\_vlm}(A_{i}|I_{i},Q_{i},P_{i}^{j}), 
\end{equation}

After training, we use $\mathcal{M}_{kc\_vlm}$ to condense the retrieved knowledge passages into concrete knowledge concepts $\bm{C}_{i}$ for each sample as follows:
\begin{equation}
    \label{eq:kc_generate}
    C_{i}^{j} = \arg \max_{y} \mathcal{M}_{kc\_vlm}(y|I_{i},Q_{i},P_{i}^{j}),
\end{equation}
where $C_{i}^{j}$ denotes the knowledge concept generated with $j$-th knowledge of the $i$-th sample. Thus, we can get $K$ knowledge concepts $\bm{C} = \{C^1, C^2, \dots, C^k\}$ from $k$ retrieved knowledge for each sample. We use the beam search to generate the knowledge concepts during the inference phase. The knowledge concept $C_{i}^{j}$ is one or two words in our experiments.

\vspace{0.5em}
\noindent\textbf{Large Language Model.} 
Recent large language models have demonstrated powerful language understanding capabilities. Many KB-VQA methods \cite{lin2022revive, gui2021kat} treat LLM as a world knowledge source to generate higher-quality implicit knowledge. 
Different from them, we treat LLM as a summarizer to extract useful information from external retrieved knowledge massages. Due to the high cost of GPT-3, we adopt the open-source large language model Vicuna \cite{vicuna} as the knowledge condenser. We first use a captioner \cite{wang2022ofa} to obtain the visual context $V$ for each image of the sample, providing the visual information to the LLM. Then we prompt the frozen LLM to select the concise information of each retrieved knowledge passage and predict the answer of the question according to the caption and the knowledge. The prompt template can be found in the Appendix. We generate the knowledge essence $\bm{E} = \{E^1, E^2, \dots, E^k\}$ from $k$ retrieved knowledge passages for each sample as follows:
\begin{equation}
    \label{eq:kc_generate}
    E_{i}^{j} = \arg \max_{y} \mathcal{M}_{kc\_llm}(y| \text{Template}(V_{i},Q_{i},P_{i}^{j})),
\end{equation}



\begin{figure}[t]
  \centering
   \includegraphics[width=1.0\linewidth]{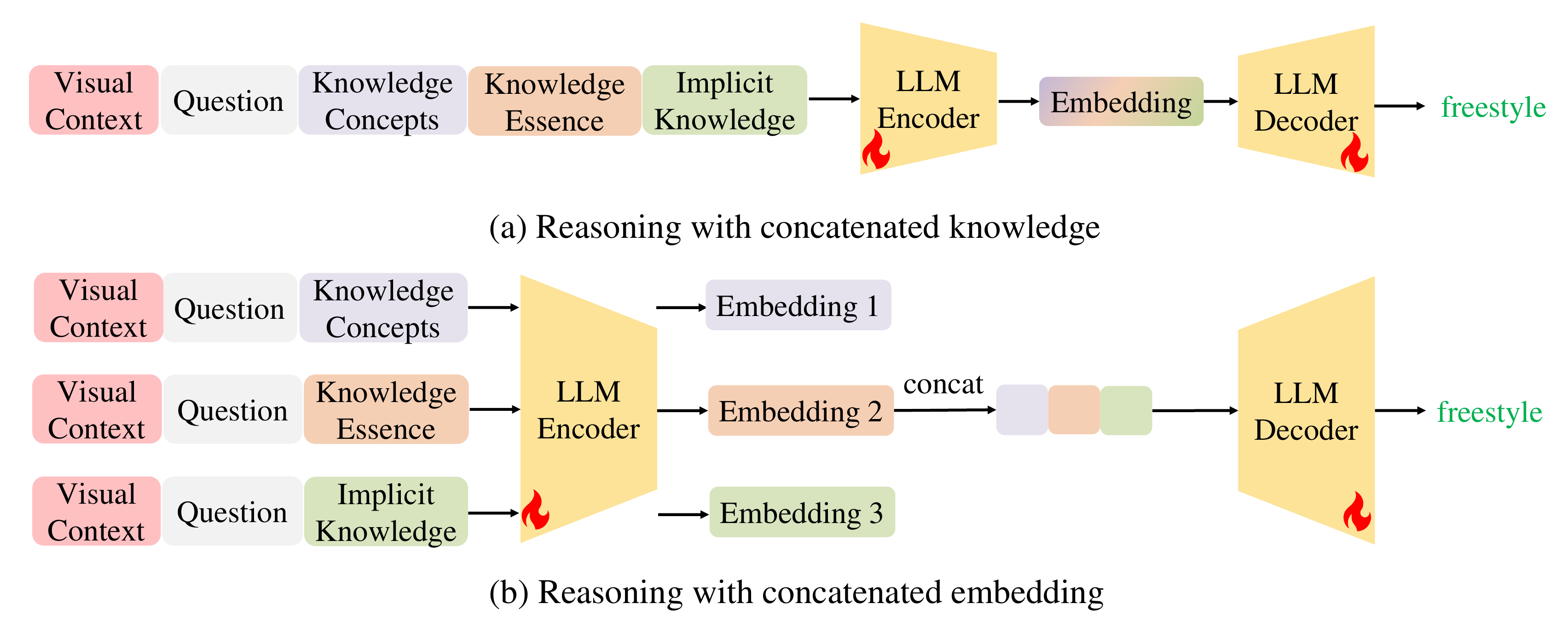}
   \caption{\textbf{The structure of the knowledge reasoning model.} (a) We concatenate the visual context, question, the condensed knowledge concepts and essence, and the implicit knowledge as a sentence and encode these information, then the decoder generates the final answer. (b) We concatenate the visual context and question with different types of knowledge as different sentences. We encode these sentences into different embeddings and they are concatenated into the decoder for generating the answer.}
   \label{fig:krmodel}
   \vspace{-1em}
\end{figure}

\subsection{Knowledge Reasoning Model} 
\label{kr}



After obtaining the condensed knowledge concepts $\bm{C}$ and the knowledge essence $\bm{E}$, we utilize an encoder-decoder architecture to jointly reason over these knowledge to select proper information to predict the answer. Further enhancing our model, we incorporate insights from the existing work Prophet \cite{shao2023prompting}. Our model integrates the output from the established Visual Question Answering (VQA) model MCAN \cite{yu2019deep} as an auxiliary knowledge source, denoted $\mathcal{O}$. This strategic integration is intended to bolster the comprehension and reasoning faculties of the model.

Given a visual context $V$, a question $Q$, a set of knowledge concepts $\bm{C}$, a set of knowledge essence $\bm{E}$ and implicit knowledge $\mathcal{O}$, we adopt two types of knowledge reasoning methods to generate the answer (as shown in Fig.~\ref{fig:krmodel}).

\vspace{0.5em}
\noindent\textbf{Concatenated knowledge.} We concatenate the visual context, question, the condensed knowledge concepts and essence, and the implicit knowledge as one sentence, then the sentence is fed into a sequence of encoder layers to jointly encode these textual information. Then we get the embedding $X^{concat} \in R^{m\times d}$, where $m$ is the number of tokens and $d$ is the embedding dimension. We pass the embedding to a sequence of decoder layers for answer generation. The model is trained under the supervision of the ground truth answer $A_{i}$:
\begin{align} \label{eq:kr_loss}
    \bm{L}_{kr} = - \sum_{i=1}^N \sum_{j=1}^{K} \log \mathcal{M}_{kr}(A_{i}|V_{i},Q_{i},P_{i}^{j}), 
\end{align}

During the inference stage, we use the beam search to generate the final answer. 

\vspace{0.5em}
\noindent\textbf{Concatenated embedding.} We concatenate the visual context and question with different types of knowledge as different sentences. These sentences are fed into a sequence of encoder layers to separately encode different information. Then we  concatenate these embeddings into a global representation $X^{multi} \in R^{3m\times d}$ and pass them into the decoder. The model training and inference are as same as above.  




\section{Experiments}
\label{sec:experiments}

\subsection{Experimental Settings}
\noindent\textbf{Dataset}. We mainly evaluate the performance of our model on OK-VQA \cite{marino2019ok} and A-OKVQA \cite{Schwenk2022AOKVQAAB} datasets. \textbf{OK-VQA} \cite{marino2019ok} is a challenging VQA dataset where a large portion of questions needs external knowledge to answer, which needs the model to have a stronger reasoning ability. The dataset contains 9K and 5K image-question pairs for training and testing, respectively. \textbf{A-OKVQA} \cite{Schwenk2022AOKVQAAB} is an augmented successor of OK-VQA, containing 25K image-question pairs that require broader commonsense and world knowledge
to answer.
No extra knowledge resources are provided for these datasets, so the researcher needs to find external world knowledge to help answer the question. 
In this paper, we use Google Search Corpus \cite{luo2021weakly} as the knowledge base for OK-VQA. It is collected in the websites using the Google Search API and covering the most knowledge which questions demand. Previous studies \cite{luo2021weakly} have proven the effectiveness of the knowledge source on the OK-VQA dataset. As for A-OKVQA, we use Wikipedia \cite{vrandevcic2014wikidata} as the knowledge base. We do not evaluate our method on traditional KB-VQA datasets \cite{wang2017fvqa, cao2021knowledge}, as they emphasizes utilizing knowledge triplets in the knowledge graph to reason the final answer rather than utilizing open-world knowledge passages. Our method focus on condensing the retrieved world knowledge to answer open questions.

\vspace{0.5em}
\noindent\textbf{Evaluation Metric.} In our experiments, we use standard VQA
evaluation metric \cite{marino2019ok} to evaluate the performance of our method. The VQA score is calculated as follows:
\begin{equation} \label{eq:1}
    \text{VQAScore}(y, \bm{A}) = \min (\frac{\#a(y)}{3}, 1)
\end{equation}
where $\bm{A}$ is a set of ground-truth answers annotated by humans, $\#a(y)$ is the  occurrence of $y$ in $\bm{A}$.

\vspace{0.5em}
\noindent\textbf{Implementation Details.} In our experiments, we adopt BLIP2 T5-XXL (11B) and Vinuna (7B) as knowledge condensation models. And we initialize our knowledge reasoning model with the pre-trained T5-XL (3B) model. We only train the Q-Former of BLIP2 and T5-XL in this paper. We use 2 Nvidia A800 GPUs (80G) for all experiments. We use Adam as the optimizer and the global batch size is set to 8 (for both models). When training the VLM of the knowledge condensation model, we use learning rate of 1e-5 and utilizes a cosine annealing learning strategy with an initial learning rate 1e-5 and a final learning rate 0 after 10 epochs. When training the knowledge reasoning model, we use learning rate of 8e-5 and utilizes a cosine annealing learning strategy with an initial learning rate 8e-5 and a final learning rate 0 after 10 epochs.
Similar to previous methods \cite{lin2022retrieval, luo2021weakly}, we use the retrieved TOP-5 knowledge passages by DPR \cite{karpukhin2020dense} from the Google Search Corpus \cite{luo2021weakly} for OK-VQA. For A-OKVQA, we use the retrieved knowledge passages by DPR \cite{karpukhin2020dense} from Wikipedia \cite{vrandevcic2014wikidata}.
We will release our codes upon paper acceptance. 

\begin{table}[t]
  \centering
  \caption{\textbf{Performance comparison with state-of-the-art (SOTA) methods on the OK-VQA dataset.}  Knowledge Sources: \textbf{C}onceptNet (C); \textbf{W}ikipedia (W); \textbf{G}oogle \textbf{S}earch (GS); \textbf{G}oogle \textbf{I}mages (GI). The best result in the table is bolded. Compared with the previous SOTA method RA-VQA-v2 \cite{Lin2023FinegrainedLM}, our method achieves 65.1\% performance, with a significant performance improvement 3.0\%.}
  \resizebox{\linewidth}{!}{
  \begin{tabular}{l|c|c|c}
    \toprule
    Method &  Large Models & Knowledge Resource & Accuracy (\%) \\
    \midrule
    \textbf{In-Context Learning \& Zero-Shot}\\
    PICa \cite{yang2022empirical}  & GPT-3 (175B)& GPT-3 (175B) & 48.0\\
    Flamingo \cite{alayrac2022flamingo}  & Flamingo (80B) & Flamingo (80B)&57.8\\
    PromptCap \cite{hu2022promptcap}  & GPT-3 (175B) & GPT-3 (175B) & 60.4\\
    Prophet \cite{shao2023prompting} & GPT-3 (175B) & GPT-3 (175B)+MCAN &61.1\\
    \midrule
    \textbf{End-to-End Finetuning} \\
    BAN+AN \cite{marino2019ok} & - & W & 25.6\\
    ConceptBERT \cite{garderes2020conceptbert} & - & C & 33.7 \\
    KRISP \cite{marino2021krisp}  &-&C+W&38.4\\
    Visual Retriever-Reader \cite{luo2021weakly}  & - & GS&39.2\\
    MAVEx \cite{wu2022multi}  & - & W+C + GI & 39.4\\
    TRiG(Ensemble) \cite{gao2022transform} & T5-large (770M) & W & 50.5 \\
    KAT(Single) \cite{gui2021kat} & T5-large (770M) & W + GPT-3 (175B) & 53.1 \\
    KAT(Ensemble) \cite{gui2021kat} & T5-large (770M) & W + GPT-3 (175B) & 54.4 \\
    RA-VQA \cite{lin2022retrieval}  & T5-large (770M) & GS & 54.5\\
    REVIVE(Single) \cite{lin2022revive}  & T5-large (770M) & W+GPT-3 (175B) & 56.6\\
    REVIVE(Ensemble) \cite{lin2022revive}  & T5-large (770M) & W+GPT-3 (175B) & 58.0\\
    BLIP2 T5-XXL \cite{Dai2023InstructBLIPTG}  & BLIP2 T5-XXL (11B) & BLIP2 & 54.7\\
    RA-VQA-v2 \cite{Lin2023FinegrainedLM} & BLIP2 T5-XL (3B)& GS &62.1\\
    \midrule
    Ours& T5-large (770M) & GS + MCAN & 63.09\\
    \textbf{Ours}  & T5-XL (3B) & GS + MCAN & \textbf{65.1} \\
    \bottomrule
  \end{tabular}}
  \label{tab:sota}
  \vspace{-0.5em}
\end{table}

\begin{table}[h]
  \centering
  \caption{\textbf{Performance comparison with state-of-the-art (SOTA) methods on the
A-OKVQA dataset.}}
  \begin{tabular}{c|cc|cc}
    \toprule
    \multirow{2}*{Method}& \multicolumn{2}{c|}{Multiple Choice} & \multicolumn{2}{c}
    {Direct Answer}\\
    ~ & val& test &val& test\\
    \hline
    ClipCap \cite{Schwenk2022AOKVQAAB} & 44.0 & 43.8 &18.1 &15.8 \\
    Pythia \cite{jiang2018pythia} & 49.0 &40.1 &25.2 &21.9 \\
    ViLBERT \cite{lu2019vilbert}& 49.1 &41.5 &30.6 &25.9 \\
    LXMERT \cite{tan2019lxmert}& 51.4 &41.6 &30.7 &25.9 \\
    KRISP \cite{marino2021krisp}& 51.9 &42.2 &33.7 &27.1 \\
    GPV-2 \cite{kamath2022webly}& 60.3 &53.7 &48.6 &40.7 \\
    PromptCap + GPT-3 \cite{hu2022promptcap}&73.2 &73.1 &56.3 &59.6 \\
    \hline
    \textbf{Ours} &\textbf{76.2}& \textbf{75.4}& \textbf{58.1}& \textbf{60.1}\\
    \bottomrule
  \end{tabular}
  \label{tab:aok}
  \vspace{-0.5cm}
\end{table}
\subsection{Comparison with State-of-the-art Methods}
\noindent\textbf{Results on OK-VQA.}
We compare the performance of our method with the state-of-the-art KB-VQA models on OK-VQA. The results are shown in Tab.~\ref{tab:sota}. According to the results, we can conclude that our method outperforms all existing methods on OK-VQA and achieves 65.1\% accuracy. 
Early models (BAN+AN \cite{marino2019ok}, ConceptBERT \cite{garderes2020conceptbert}, KRISP \cite{marino2021krisp}, Visual Retriever-Reader \cite{luo2021weakly}, and MAVEx \cite{wu2022multi}) uitilize small visual-language backbones and integrate external knowledge to generate the answer. 
Although these methods use various sources of knowledge bases such as  ConceptNet, Wikipedia, Google Web Search and Google Images, they achieve relatively low VQA score on OK-VQA from 25.6\% to 39.4\% accuracy. 
Due to a vast knowledge and a great capacity for common-sense reasoning,  large language models are utilized to improve the performance on OK-VQA. Some studies (TRiG\cite{gao2022transform}, KAT \cite{gui2021kat}, RA-VQA \cite{lin2022retrieval}, and REVIVE \cite{lin2022revive}) exploit T5-large to question answering in conjunction with external knowledge, including explicit and implicit knowledge. 
Compared with early methods, they greatly improve the performance of OK-VQA with accuracy ranging from 50.5\% to 58.0\%. 
Recently, PICa \cite{yang2022empirical} first prompts the large-scale language model GPT-3 to generate the answer and achieves promising performance with in-context learning. Inspired by the PICa success, some methods \cite{hu2022promptcap, shao2023prompting} improve the reasoning ability of GPT-3 by introducing question-guided captions and providing answer candidates, achieving higher VQA score on OK-VQA, about 61\%. 
Meanwhile, KAT \cite{gui2021kat} and REVIVE \cite{lin2022revive} consider GPT-3 as the implicit knowledge base and prompt the model to generate some candidates and corresponding explanations, which obtain significant performances compared with methods only using the explicit knowledge base. 
Except for these LLM-based methods, large visual-language models have achieved high performance on OK-VQA without using external knowledge(\emph{e.g.}, Flamingo \cite{alayrac2022flamingo} (80B) has a few-shot VQA score of 57.8\% and BLIP2 T5-XXL has a finetuning VQA score of 54.7\%). Based on large visual-language models, RA-VQA-v2 \cite{Lin2023FinegrainedLM} finetunes BLIP2 T5-XL with external knowledge and achieves a VQA score of 62.1\%.

We propose the knowledge condensation model to distill concise knowledge concepts and essence from verbose knowledge passages, filtering irrelevant and distractive information. Then these condensed knowledge concepts and essence are passed into the knowledge reasoning model for KB-VQA task. Therefore, our method outperforms all state-of-the-art with large margins, achieving a VQA score of 65.1\% with a relatively small model size (3B). Even with smaller model size (770M), our model still achieves the SOTA performance of 63.09\%.
The significant performance improvement demonstrates the effectiveness of our method.

\vspace{0.5em}
\noindent\textbf{Results on A-OKVQA.} Recent studies based on external knowledge retrieving only conduct the experiments on OK-VQA, it has not been proven that the effectiveness of using external knowledge. In this paper, we use Wikipedia \cite{vrandevcic2014wikidata} as the external knowledge base and use a pre-trained DPR to retrieve the knowledge passages. Except the knowledge base, the experiment settings are the same as OK-VQA. 
Tab.~\ref{tab:aok} shows the comparative results on the challenging A-OKVQA dataset. The results show that our method surpasses previous state-of-the-art methods in both Direct Answer and Multiple Choice settings, which demonstrates the effectiveness and generalization of our method. 

\subsection{Ablation Studies}

We conduct the ablation studies on OK-VQA to evaluate the key components of our method.

\begin{table}[t]
  \centering
  \caption{ \textbf{Ablation study on the condensation model of our method on OK-VQA dataset.} The best results in the table are shown in bold. \textbf{Knowledge Passages} denote the passages retrieved by DRR. \textbf{Implicit Knowledge} denotes the answer candidates generated by the pre-trained VQA model MCAN \cite{karpukhin2020dense}. \textbf{knowledge essence} denotes the condensed knowledge by the LLM of the condensation model. \textbf{Knowledge Concepts} denotes the condensed knowledge by the VLM of the condensation model.}
  \begin{tabular}{c|c|c|c|c|c}
    \toprule
    Experiments & \makecell{Knowledge \\ Passages}  &  \makecell{Knowledge \\ Essence}& \makecell{Knowledge \\ Concepts} & \makecell{Implicit \\ Knowledge} &Accuracy (\%) \\
    \midrule
    1&& & & &55.29 \\
    2&$\checkmark$ &  & &  &58.69\\
    3&& $\checkmark$ &  & &60.77\\
    4&&  & $\checkmark$ & &62.26\\
    \midrule
    5&&  & & $\checkmark$ &58.69 \\
    6&$\checkmark$ &  &  & $\checkmark$ &61.28\\
    7& & $\checkmark$ &  & $\checkmark$&62.77\\
    8& &  & $\checkmark$ & $\checkmark$ &64.53\\
    
     \midrule
     9& & $\checkmark$  & $\checkmark$ &$\checkmark$& \textbf{65.10}\\
     10& $\checkmark$ & $\checkmark$  & $\checkmark$ &$\checkmark$& 64.90\\
    \bottomrule
  \end{tabular}
  \label{tab:condense}
  \vspace{-1em}
\end{table}

\noindent\textbf{Effect of the knowledge condensation model.} We conduct detailed experiments to show the effectiveness of the knowledge condensation model. As shown in Tab.~\ref{tab:condense}, we have the following observations: 
\begin{enumerate}
\item Compared to the baseline method without using any external knowledge (the 1st line), introducing retrieved knowledge passages (the 2nd line) improves the performance from 55.29\% to 58.69\%. When replacing the knowledge passages with our condensed knowledge essence and concepts (the 3rd and 4-th line), the performance improves from 58.69\% to 60.77\%, and 62.26\%, respectively. It turns out that both types of condensed knowledge help to reason out the correct answer.

\item Compared to the method with knowledge passages and implicit knowledge (the 6-th line), replacing verbose passages with concise knowledge essence and concepts (the 7-th and 8-th line) brings 1.39\% and 3.25\% performance improvement, achieving 62.77\% and 64.53\% on OK-VQA dataset. 
The results demonstrate that both types of condensed knowledge retain more valid information than lengthy knowledge passages.

\item Compared to the method with only implicit knowledge (the 5-th line), combining the knowledge essence and concepts with the implicit knowledge (the 9-th line) brings 6.41\% performance improvement and achieves SOTA performance \textbf{65.10}\%. This fact verifies that the knowledge essence and concepts can complement each other.

\item It is worth noting that adding knowledge passages (the 10th line) to experiments with knowledge essence, knowledge concepts, and implicit knowledge (the 9th line) leads to marginal performance reduction. We argue that condensed knowledge essence and concepts have absorbed effective information from the original knowledge passages, while irrelevant information in knowledge passages may disturb the model reasoning.
\end{enumerate}
All these comparisons verify the effectiveness of the knowledge condensation model under different experimental settings.

\vspace{0.5em}
\noindent\textbf{Effect of the number of knowledge essence and knowledge concepts.} 
In this part, we investigate the effect of different numbers of knowledge essence and knowledge concepts on performance. In the knowledge essence (concept) experiment, the number of knowledge concepts (essence) is set to 5. As the number of knowledge concepts and essences increases, the model performance gradually improves. This indicates that the more valid information in the knowledge, the more it helps the model to reason out the correct answer.


\begin{table}[t]
  \centering
  \begin{minipage}[t]{.55\linewidth}
  \centering
  \caption{Ablation study on the number of condensed knowledge.}
  \vspace{-1em}
  \resizebox{0.7\linewidth}{!}{
  \begin{tabular}{c|c|c}
    \toprule
     \multirow{2}{*}{Number}& \multicolumn{2}{c}{Accuracy (\%)}\\
     \cmidrule(r){2-3}
       & Knowledge Essence & Knowledge Concepts \\
    \midrule
     1 & 64.00& 64.32\\
     2 & 64.58& 64.49\\
     3 & 64.84& 64.74\\
     4 & 65.06& 64.92\\
     5 & 65.10& 65.10\\
    \bottomrule
  \end{tabular}}
  \label{tab:num}
  \end{minipage}
  \hfill
  \begin{minipage}[t]{.4\linewidth}
  \centering
  \caption{Ablation study on different VLMs of knowledge condensation model.}
  \resizebox{\linewidth}{!}{
  \small
  \begin{tabular}{c|c|c}
    \toprule
     BLIP2 T5-XL  & BLIP2 T5-XXL &Accuracy (\%) \\
    \midrule
    $\checkmark$ &  &64.01\\
     & $\checkmark$ &65.10\\
    $\checkmark$ & $\checkmark$ &65.51\\
    \bottomrule
  \end{tabular}}
  \label{tab:vlms}
  \end{minipage}
  \vspace{-1em}
\end{table}

\vspace{0.5em}
\noindent\textbf{Effect of different VLMs in the knowledge condensation model.} 
In this study, different VLMs are adopted to condense the knowledge passages into knowledge concepts. While maintaining other modal inputs at their default configurations, we extend our investigation to include the BLIP2 T5-XL model alongside the BLIP2 T5-XXL. As shown in Tab.~\ref{tab:vlms}, we find that condensing with stronger VLM leads to a higher performance. The ensemble of knowledge concepts by different VLMs can bring performance improvement. This finding underscores the utility of leveraging a variety of VLMs to generate knowledge concepts, which in turn provides a richer and more complementary set of information for KB-VQA tasks.


\vspace{0.5em}
\noindent\textbf{Effect of different backbones of knowledge reasoning model.}
In this part, different sizes of T5 \cite{chung2022scaling} model are adopted as the knowledge reasoning model. As shown in Tab.~\ref{tab:different t5}, regardless of the type of knowledge used, increasing model size from T5-base (250M), T5-large (770M) to T5-XL (3B) can lead to model performance improvement. On the other hand, the knowledge reasoning model based on T5-base (250M) with knowledge essence and concepts achieves a comparable performance of 60.92\% compared to the counterpart of T5-XL (3B) with knowledge passages 61.28\%. These results demonstrate the effectiveness of our condensed knowledge essence and concepts.


\vspace{0.5em}
\noindent\textbf{Effect of different reasoning methods of knowledge reasoning model.} We evaluate the two proposed knowledge reasoning methods described in Fig.~\ref{fig:krmodel} under the different backbone settings. From Tab.~\ref{tab:different reason}, we find that the knowledge reasoning method with concatenated embedding achieves comparable performance to the counterpart with concatenated knowledge. Taking T5-base and T5-large as the backbone, the knowledge reasoning method with concatenated embedding achieves slightly higher performance. Instead, the knowledge reasoning method with concatenated knowledge achieves slightly higher performance on the T5-XL backbone. Thus, we adopt the knowledge reasoning method with concatenated knowledge and T5-XL as the default setting.

\begin{table}[t]
\begin{minipage}[t]{0.5\linewidth}
  \centering
  \caption{Ablation study on different backbones of knowledge reasoning model.} 
  \resizebox{1\linewidth}{!}{
  \begin{tabular}{c|c|c}
    \toprule
    Types of Knowledge  & Backbone & Accuracy (\%)\\
    \midrule
    \multirow{3}{*}{\makecell{Knowledge \\Passages}}& T5-base& 55.87\\
       &T5-large & 56.25\\
       &T5-XL & 61.28\\
    \midrule
    \multirow{3}{*}{\makecell{Knowledge \\Essence\&Concepts}}& T5-base& 60.92\\
       &T5-large & 63.09\\
    &T5-XL & 65.10\\
    \bottomrule
  \end{tabular}}
  \label{tab:different t5}
\end{minipage}
\hfill
\begin{minipage}[t]{0.5\linewidth}
  \centering
  \caption{Ablation study on different knowledge reasoning methods.} 
  \resizebox{0.95\linewidth}{!}{
  \begin{tabular}{c|c|c}
    \toprule
    Reasoning Method  & Backbone & Accuracy (\%)\\
    \midrule
    \multirow{3}{*}{\makecell{Concatenated \\knowledge}}& T5-base& 60.92\\
       &T5-large & 63.09\\
       &T5-XL & 65.10\\
    \midrule
    \multirow{3}{*}{\makecell{Concatenated \\embedding}}& T5-base& 61.22\\
       &T5-large & 63.19\\
    &T5-XL & 64.65\\
    \bottomrule
  \end{tabular}}
  \label{tab:different reason}
\end{minipage}
\vspace{-1em}
\end{table}

\begin{wraptable}{r}{0.5\linewidth}
  \centering
  \vspace{-3em}
  \caption{Ablation study on different visual information of the knowledge reasoning model.}
  \begin{tabular}{c|c}
    \toprule
    Visual Information &Accuracy (\%)\\
    \midrule
    Feature token & 65.06\\
    Caption  & 65.10 \\
    Caption + Feature token& 65.19\\
    \bottomrule
  \end{tabular}
  \label{tab:different visual}
  \vspace{-2em}
\end{wraptable}
\vspace{0.5em}
\noindent\textbf{Effect of different visual information of knowledge reasoning model.} We explore the effect of visual information in the knowledge reasoning model. Feature token in Tab.~\ref{tab:different visual} indicates the image feature tokens extracted by the pretrained Q-Former of BLIP2 T5-XL. Captions are obtained following the previous methods \cite{lin2022revive, gui2021kat, lin2022retrieval}.
As shown in Tab.~\ref{tab:different visual}, compared to the image caption, adopting visual tokens from the BLIP2 Q-Former cannot bring performance improvement. We argue that the visual image has been utilized in the knowledge condensation model to generate the knowledge concepts. As a result, introducing the visual tokens based on the knowledge concepts and essence cannot bring performance improvement.

\begin{figure*}[t]
  \centering
   \includegraphics[width=0.98\linewidth]{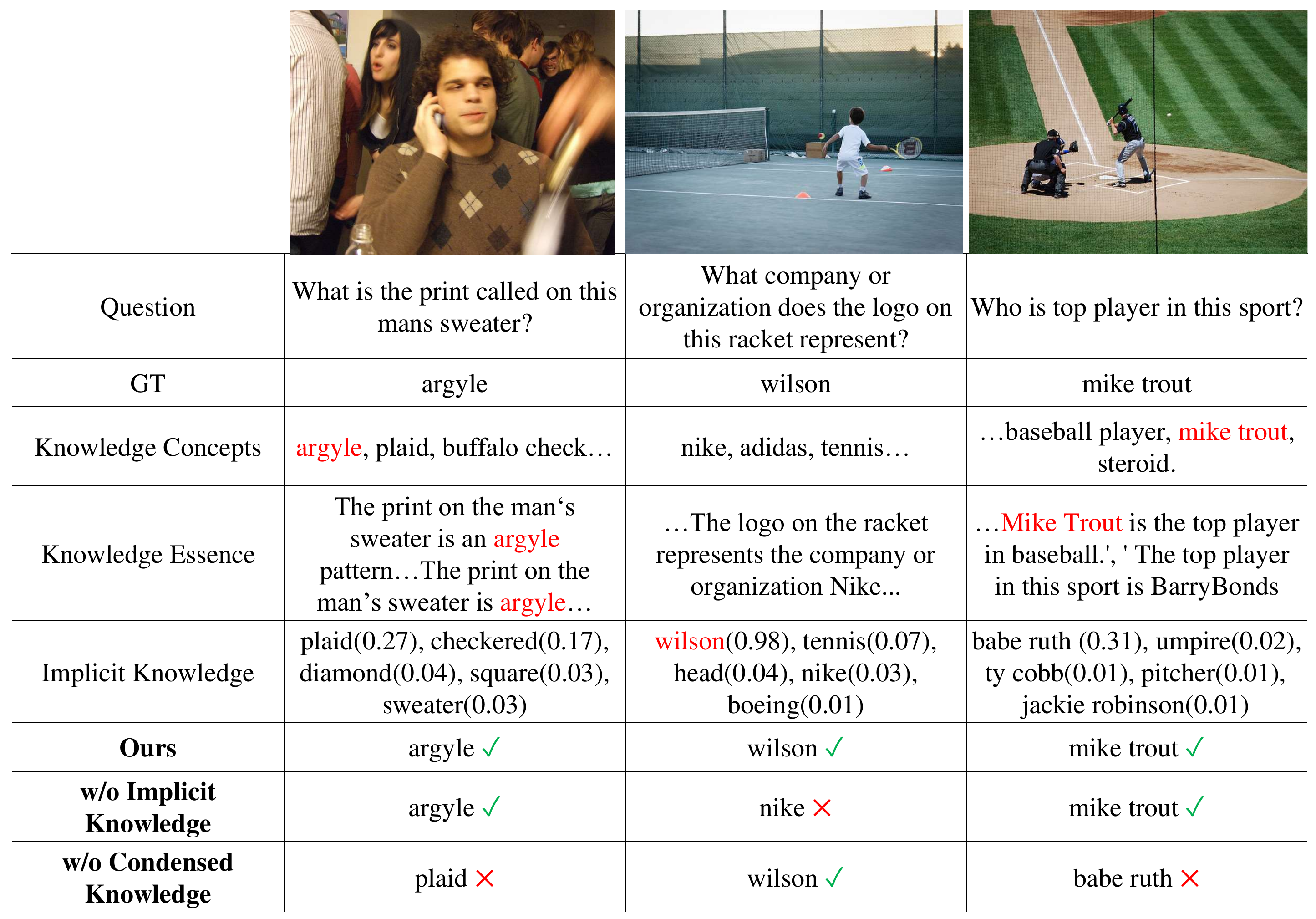}
   \caption{\textbf{Case study of our method.} The condensed knowledge are distilled valid information from retrieved knowledge passages by the knowledge condensation model. The implicit knowledge and corresponding scores (in brackets) are produced by a pre-trained VQA classification model MCAN.  
   The condensed knowledge can provide key information to help the model to reason the answer. By introducing extra implicit knowledge, the knowledge reasoning model can reason over more knowledge bases to select the right knowledge for answering the question.}
   \label{fig:case}
   \vspace{-0.5em}
\end{figure*}



\subsection{Case study} 
As shown in Fig.~\ref{fig:case}, we further analyze the effectiveness of different components of our method. The first row shows that: when using the knowledge concepts and essence condensed by the knowledge condensation model, the model predicts the right answer "argyle". Instead, the model using original retrieved knowledge passages gives the wrong answer "plaid". This means that the distilled knowledge from original noisy knowledge passages can extract the valid information to the reasoning model eliminating interference information. This reduces the difficulty of reasoning for the model.
The second row shows the complementary of the implicit knowledge and our post-processing knowledge. Sometimes the vanilla VQA model can provide extra useful information using the implicit reasoning ability. In this case, the condensation model transforms all the knowledge passages into useless information probably because original knowledge passages contain too much distractive information. The implicit knowledge helps the model answer the right answer "wilson". 
The third row shows the strong information utilization ability of the knowledge reasoning model. Although the right answer is not the first order in the knowledge concepts, the model can reason the right answer without adopting the first-order information in the condensed knowledge and the implicit knowledge. 

\section{Conclusion}
In this paper, we propose the knowledge condensation model and the knowledge reasoning model to utilize the retrieved external knowledge effectively. We leverage the multimodal perception and reasoning ability of VLM and the text understanding ability of LLM to distill valid information from noisy retrieved knowledge passages. Then we propose a encoder-decoder knowledge reasoning model to utilize various information for answer generation.
Experimental results on OKVQA and A-OKVQA demonstrate the effectiveness of our method. We hope that our work can provide a new idea about how to distill the knowledge to solve the knowledge-based VQA problem.

%
%
\bibliographystyle{splncs04}
\bibliography{main}

\section{Appendix}
\subsection{Prompt Template}
For the large language model in Knowledge Condensation model, we use a pre-designed prompt template to condense each knowledge passage given a sample (shown in Tab.~\ref{tab:prompt}). 
\vspace{-0.5em}
\begin{table}[h]
	\centering
	\caption{Prompt Template used in Knowledge Condensation model.}
        \begin{tabular}{p{10cm}}
            \toprule
            Prompt Template\\
            \midrule
            Select the key information of the retrieved knowledge and 
            answer the question according to the visual context and concise knowledge.
            Let's think step by step. \\
            Visual context: \textcolor{blue}{$V$} \\
            Retrieved knowledge: \textcolor{blue}{$P^j$} \\
            ===\\
            Question: \textcolor{blue}{$Q$} \\
            Answer: \\
    \bottomrule
\end{tabular}
\vspace{-0.5em}
\label{tab:prompt}
\end{table}

\subsection{More Quantitative Results}
\noindent\textbf{Effect of different number of Knowledge Passages and Knowledge Essence \&  Concepts.} Tab.~\ref{tab:num} varies the number of different types of knowledge from 0 to 5 to explore its effect on the model performance. From the results, we can see that with the increase of number, the accuracy of models using different types of knowledge grow accordingly and achieves the best VQA score when the number is set to 5. And the model using condensed knowledge essence and concepts outperforms than using retrieved knowledge passages under different size settings. The results show the effectiveness of our method.

\vspace{0.5em}
\noindent\textbf{Performance of our method on different categories.} We show the per-category accuracies of our method and the base model in Tab.~\ref{tab:type}. Our method outperforms the base model on all categories, indicating the generality of our method.

\begin{table}[t]
  \centering
  \caption{\textbf{Ablation study of different number of Knowledge Passages and Knowledge Essence \&  Concepts.} Our method achieves increasingly performance with more knowledge.}
  \begin{tabular}{c|c|c}
    \toprule
     \multirow{2}{*}{Number}& \multicolumn{2}{c}{Accuracy (\%)}\\
     \cmidrule(r){2-3}
       & \makecell{Knowledge \\Passages} & \makecell{Knowledge \\Essence \&  Concepts} \\
    \midrule
     1 & 59.24& 63.35\\
     2 & 59.96& 63.96\\
     3 & 60.34& 64.24\\
     4 & 60.77& 64.88\\
     5 & 61.28& 65.10\\
    \bottomrule
  \end{tabular}
  \label{tab:num}
\end{table}

\begin{table}[t]
  \centering
  \caption{\textbf{Per-category accuracies of the base model and our method.} Our method outperforms the base model largely on all categories.}
  \resizebox{0.8\linewidth}{!}{
  \begin{tabular}{l|c|c|c}
    \toprule
    Category & Baseline&\makecell{Baseline+\\Knowledge \\Passages} & \makecell{Baseline+\\Knowledge \\Essence \&  Concepts}\\
    \midrule
    Plants and Animals & 57.09&62.74&64.91 \\
    Science and Technology & 49.76&59.05&59.05 \\
    Sports and Recreation &56.26&64.97&68.18 \\
    Geography, History, Language and Culture &53.05&66.52&64.26 \\
    Brands, Companies and Products  &50.70&62.91&60.47 \\
    Vehicles and Transportation &50.95&58.72&62.04 \\
    Cooking and Food &58.26&64.12&67.33 \\
    Weather and Climate &57.52&69.77&71.47 \\
    People and Everyday life &56.50&61.45&64.35 \\
    Objects, Material and Clothing &56.45&58.55&61.50 \\
    \midrule
    Overall&55.29&61.28&65.10\\
    \bottomrule
  \end{tabular}}
  \label{tab:type}
\end{table}

\subsection{More Qualitative Results}
\begin{figure}[t]
  \centering
   \includegraphics[width=0.8\linewidth]{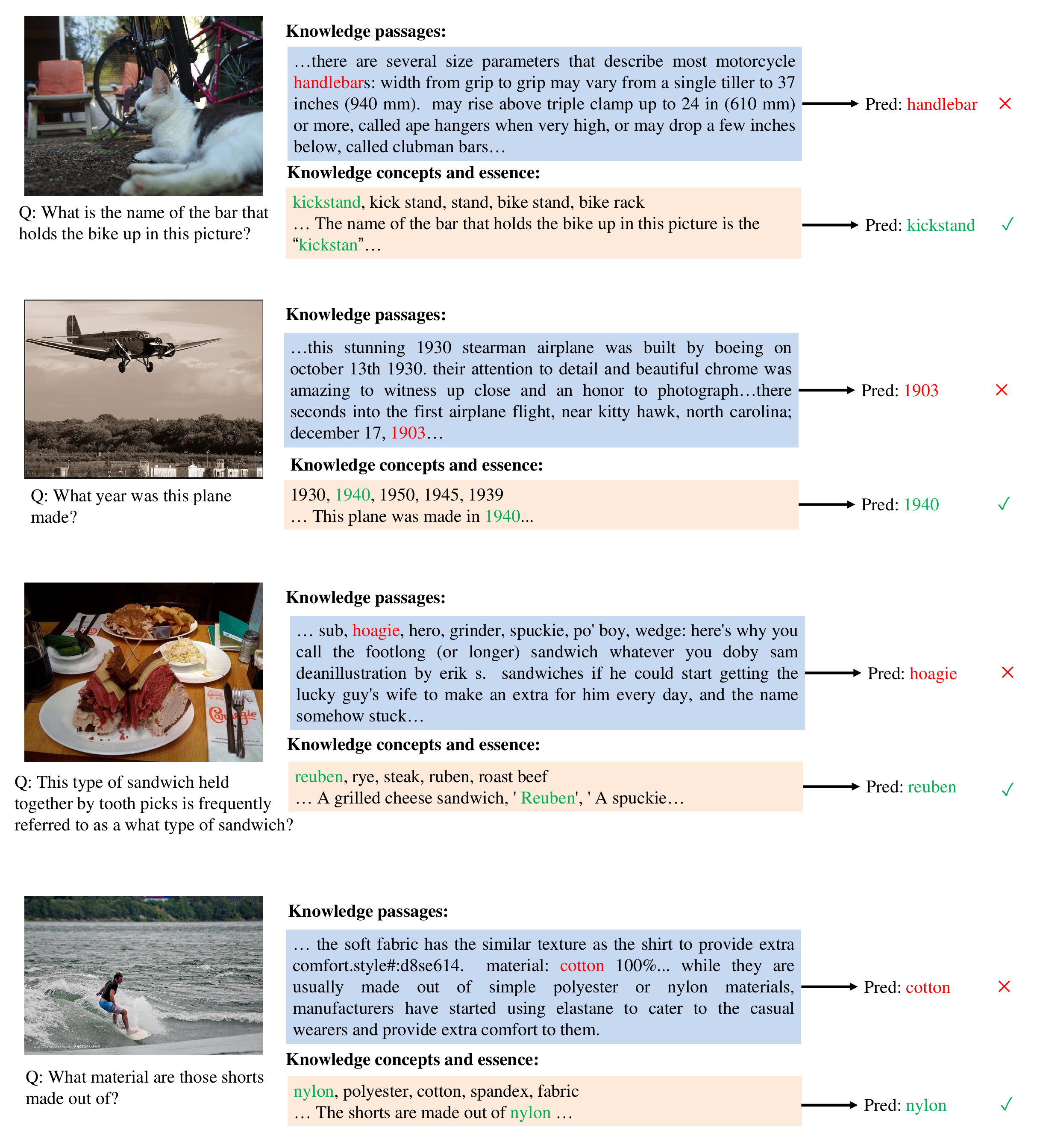}
   \caption{\textbf{Qualitative comparison between using knowledge passages and knowledge concepts\&essence.} The blue box corresponds the original retrieved knowledge passages. The yellow box corresponds the condensed knowledge concepts and essence.}
   \label{fig:case_sup1}
\end{figure}

We show the qualitative comparison of our method using knowledge passages and knowledge concepts\&essence in Fig.~\ref{fig:case_sup1}. The results show that using the condensed knowledge concepts and essence by our method can provide concise information to help the model reason the right answer. However, original knowledge passages often contain distractive information and mislead the model to predict the wrong answer.

\begin{figure}[t]
  \centering
   \includegraphics[width=0.98\linewidth]{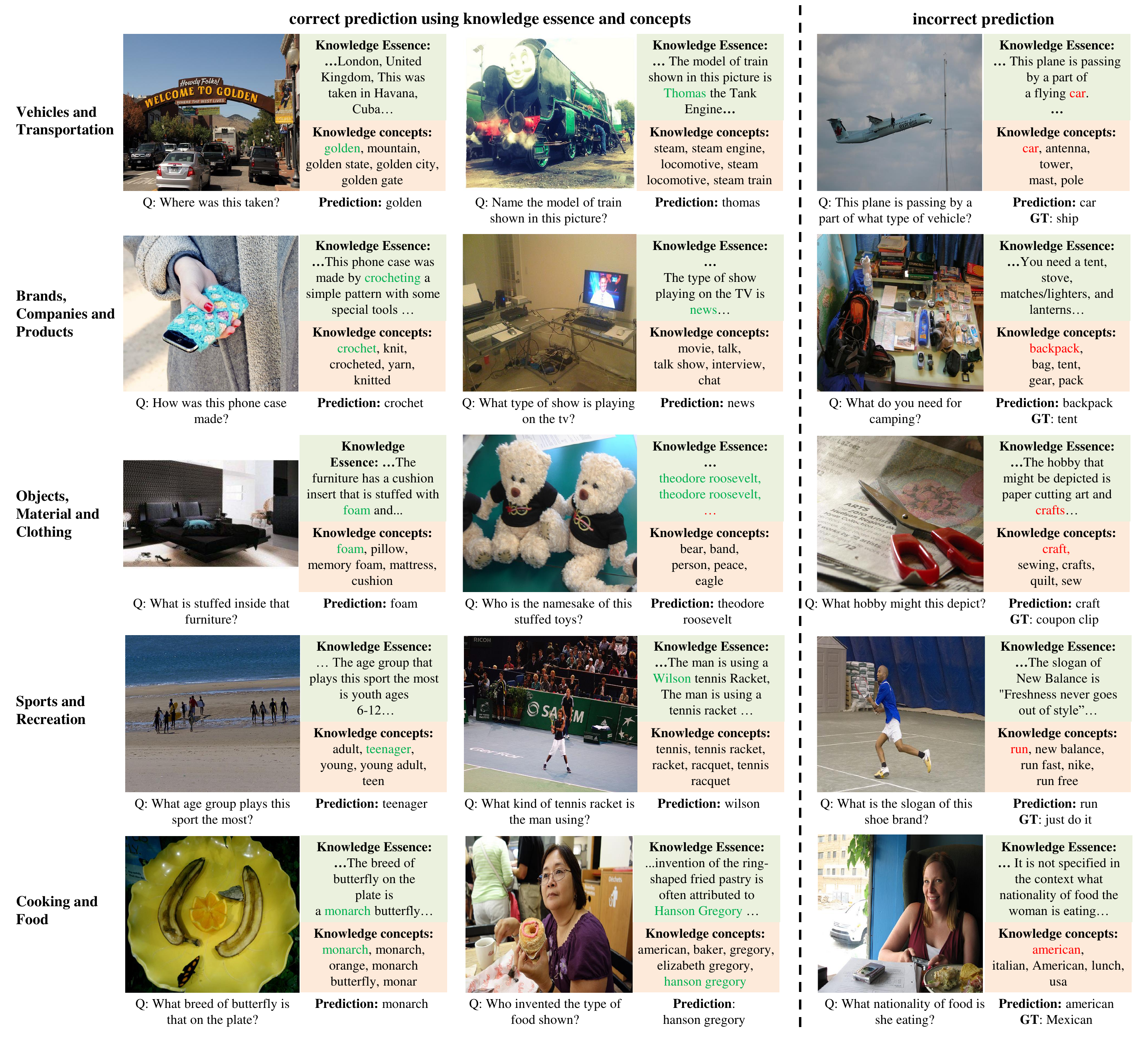}
   \caption{\textbf{Qualitative results of different categories using knowledge concepts and essence.} The first two columns correspond to the correctly answered samples using knowledge concepts and essence. The last column shows some failure samples.}
   \label{fig:case_sup2}
\end{figure}

\begin{figure}[t]
  \centering
   \includegraphics[width=0.98\linewidth]{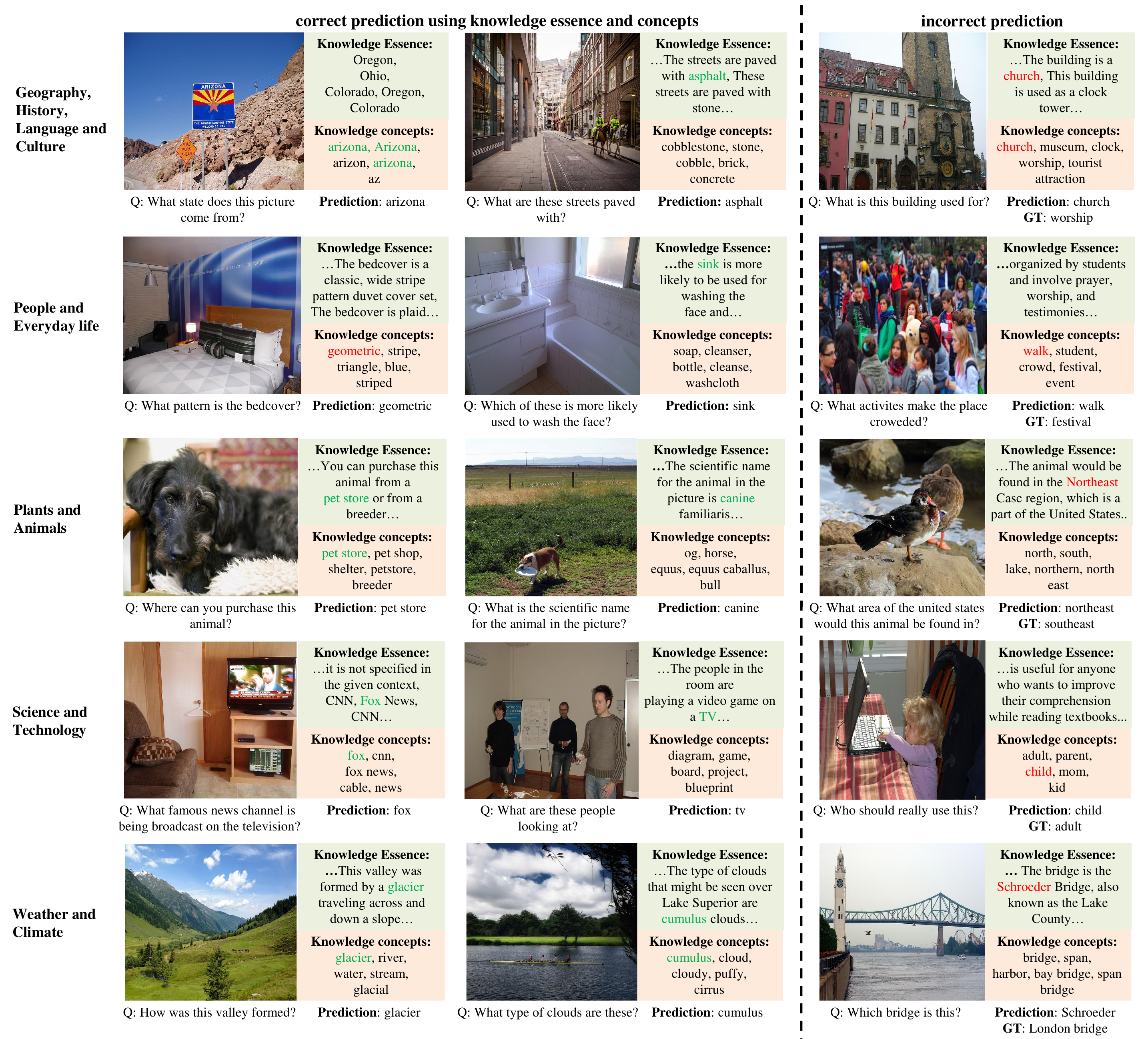}
   \caption{\textbf{Qualitative results of different categories using knowledge concepts and essence.} The first two columns correspond to the correctly answered samples using knowledge concepts and essence. The last column shows some failure samples.}
   \label{fig:case_sup3}
\end{figure}

Fig.~\ref{fig:case_sup2} and Fig.~\ref{fig:case_sup3} demonstrates some testing samples from different knowledge categories. In the first two column, we show the correctly answered samples with knowledge concepts and essence, showing the two kinds of knowledge information can provide different evidence for reasoning from different aspects. The last column shows some failure samples, implying that there is still room for future improvement.

\end{document}